\documentclass[12pt]{article}

\usepackage{amsmath}
\usepackage{amssymb}
\usepackage{float}

\usepackage{graphicx}

\usepackage{cite}

\usepackage{color} 

\usepackage[labelfont=bf,labelsep=period,justification=raggedright]{caption}

\makeatletter
\renewcommand{\@biblabel}[1]{\quad#1.}
\makeatother

\date{}

\begin{document}

\begin{flushleft}
{\Large
\textbf{A model of grassroots changes in linguistic systems}
}
\\
Janet B. Pierrehumbert$^{1,2,3\ast}$, 
Forrest Stonedahl$^{4}$, 
Robert Daland$^{5}$
\\
\bf{1} Linguistics Department, Northwestern University, Evanston, IL, USA
\\
\bf{2} Northwestern Institute on Complex Systems, Evanston, IL USA
\\
\bf{3} New Zealand Institute of Language, Brain, and Behaviour, Univ. of Canterbury, Christchurch, NZ
\\
\bf{4} Computer Science, Augustana College, Rock Island, IL, USA
\\

\bf{5} Linguistics Department, Univ. California Los Angeles, Los Angeles, CA, USA
\\
$\ast$ E-mail: jbp@northwestern.edu
\end{flushleft}

\section*{Abstract}

Linguistic norms emerge in human communities because people imitate each other. A shared linguistic system provides people with
the benefits of shared knowledge and coordinated planning. Once norms are in place, why would they ever change? This question,
echoing broad questions in the theory of social dynamics, has particular force in relation to language. By definition, an
innovator is in the minority when the innovation first occurs. In some areas of social dynamics, important minorities can strongly
influence the majority through their power, fame, or use of broadcast media. But most linguistic changes are grassroots
developments that originate with ordinary people. Here, we develop a novel model of communicative behavior in communities, and
identify a mechanism for arbitrary innovations by ordinary people to have a good chance of being widely adopted.

To imitate each other, people must form a mental representation of what other people do. Each time they speak, they must also
decide which form to produce themselves. We introduce a new decision function that enables us to smoothly explore the space
between two types of behavior: probability matching (matching the probabilities of incoming experience) and regularization
(producing some forms disproportionately often). Using Monte Carlo methods, we explore the interactions amongst the degree of
regularization, the distribution of biases in a network, and the network position of the innovator. We identify two regimes for
the widespread adoption of arbritrary innovations, viewed as informational cascades in the network. With moderate regularization
of experienced input, average people (not well-connected people) are the most likely source of successful innovations. Our results
shed light on a major outstanding puzzle in the theory of language change. The framework also holds promise for understanding the
dynamics of other social norms.

\section*{Introduction}

Many human needs and goals are shared by people in different cultures. But the social norms for
addressing them can be highly arbitrary in their details. This is especially true for 
norms that promote communication and social cohesion. The value of such norms may arise more from
the existence of a consensus than from its specifics. A fax machine has value only insofar as other
people also have compatible fax machines.  Facebook supports social contact only insofar as one's
friends and relations are also on Facebook. This situation is epitomized in the most complex and
highly evolved communication system of all, namely human language. Relationships between word forms
and word meanings are very idiosyncratic across languages, and hence must be learned for any
individual language  \cite{newmeyer, saussure}. The usefulness of language comes about because people
come to agree on ways to refer to entities, events, and abstractions. Agreement enables them to pool
individual knowledge and benefit from cooperative activities over great expanses of time and space.

Once a linguistic consensus is in place in a community,  why and how can it ever change? This
question, the central focus of our paper,  is a major challenge for models of language dynamics. It
is widely agreed that linguistic systems emerge through  innate propensities for people to imitate
others in their communities \cite{nettle, fagyal, ke, niyogi, deboer, lacerda, caldwell}. Frequent
forms, and forms produced by many different speakers, are at an advantage in being learned,
remembered, and used \cite{macwhinney, ellis, lieberman, altmann}. In contrast, infrequent forms are
vulnerable to loss over time \cite{bybee, lieberman}. These basic findings  all mitigate against
linguistic innovations, which by definition are rare and poorly disseminated when they first occur.
Nonetheless, natural languages are never perfectly stable. They change at a wide range of time
scales, and  the fastest changes (such as the adoption of new slang expressions and jargon) are
exhibited on time-scales of a few months or less within the speech of individual adults. In this
regard, linguistic systems resemble cultural norms in other domains such as fashion, dancing,
politics, and religion. In these domains as well, people are strongly influenced by the consensus of
their neighbors, but norms still evolve over time. Complex patterns of variation across groups
emerge when distinct groups follow different trajectories of cultural evolution.

Here, we develop a new model of how linguistic innovations can become widely adopted in a community.
We focus on a type of changes that are particularly challenging to model:
changes that we  characterize as  {\it fast}, {\it arbitrary}, and {\it egalitarian}.
By {\it fast} changes, we mean ones that occur through learning and adaptation by individual
speakers in the course of repeated interaction with their peers, without any generational replacement. 
Ref \cite{lenaerts} characterizes within-generation changes in social norms as cases of  {\it horizontal transmission}.
The slow changes that occur during cross-generational transmission of language fall outside of the scope of
this paper; modeling them requires distinct mechanisms \cite{niyogi, castellano}. 

By {\it arbitrary} changes, we mean changes that involve no global functional advantage. 
They may be neutral, offering no identifiable functional advantage at all.  Or, if there is
any advantage to some people from adopting them, it is balanced by some equal disadvantage to
others. We do not  deny that some changes in language may be globally advantageous. For example,
human languages are shaped by a persistent bias to encode more predictable information using shorter
expressions, which improves the efficiency of communication \cite{piantadosi}. However, prior work in
language evolution, drawing on evolutionary biology, already reveals how positive utility can cause
an innovation to spread. In contrast, for many linguistic changes, no global advantage is apparent. 
Explaining how arbitrary changes can take hold is essential to explain the vast profusion of
apparently arbitrary detail that is observed synchronically \cite{newmeyer}.

We use the term {\it egalitarian}  to highlight our interest in changes that originate with ordinary
people and spread without regard to social status.  Sociolinguistic field work has shown that people
at the forefront of linguistic change typically belong to the lower middle class or upper working
class.  Although imitation of upper-class people may play a role in some conscious choices, such as
product purchases, it  cannot be the fundamental mechanism for linguistic changes. Typically,
upper-class speakers eventually adopt patterns set by lower classes, rather than the other way
around \cite{labov2001}. This pervasive pattern could not arise if imitating the few people at the top of
the social hierarchy (perhaps in an effort to increase social capital) is the engine of linguistic
change.  However, the empirical findings do  not necessarily indicate that people preferentially
imitate lower-class people. Ordinary people are far more numerous than upper-class people, and this
factor alone might explain the results. We thus take as a starting point the null hypothesis by
which listeners weight input from all speakers equally, without assigning more value or prestige to
some speakers than to others. Unlike Refs \cite{nettle, fagyal}, we do not assume that people with
many social connections have more status than others, or are preferentially taken as linguistic
models. This null hypothesis takes us surprisingly far. We identify a confluence of social and
cognitive factors that enables fast, arbitrary, and egalitarian innovations to propagate with
considerable probability. The key cognitive factor is the decision rule relating the perception of linguistic norms to the 
choice of what form to produce on any particular occasion. The key social factor is the distribution of biases over the members of the community.
If some people are conservative, and others love novelty, the linguistic system is more likely to change than if nobody is biased in either direction.
These findings challenge previous claims \cite{nettle, fagyal, bloomfield}
that differences in social prestige are a necessary component in a theory of how linguistic innovations can
spread. 

Our model of how innovations propagate integrates 
technical ingredients from work in several disciplines.  Following recent work in network
theory, we consider the linguistic community to be a social network in which nodes represent people
and links represent social relationships. Specifically, a link  between two nodes means that the two
communicate with each other and are disposed to imitate each other
in their choices of expression.  Social ties that do not involve this level of mutual influence
exist of course, but these ties are omitted from the model.

Next, we draw a careful distinction between the mental state of a node and the signals that the node
emits. We assume that signals are categorical. For example, on each individual occasion when some
speaker wishes to quote someone else, she either uses the conservative form (e.g. {\it he said}) or
an innovative form (e.g. {\it he was like}).  Mental states, in contrast, are numbers on a continuous scale of probabilities,
representing the speaker's current estimate of the linguistic norm. They take values from $0$ (``I
don't believe anybody uses that expression'') to $1$ (``I believe everybody uses that expression''). The
value $0$ will be used here for the conservative norm, and $1$ for the innovation, so that
in this example, a mental state of $0.6$ would represent the speaker's belief that people in general
use {\it  like}  to express quotation with $P = 0.6$.  The model is applicable to any linguistic
innovation that can be described as a new category competing with a pre-existing one. Examples
include novel slang, such as {\it lol} in competition with {\it ha-ha}; novel affixes such as  {\it
uber-} in competition with {\it super-}, as well as new constructions, such as {\it gonna} in
competition with {\it will} for expressing the future. It does not encompass gradient changes in the
signals themselves, such as the gradual shifts in vowel quality explored in Ref \cite{sonderegger}.

Signals and mental states are linked by a decision function, whose input is the mental state and
whose output is the probability of emitting different signals. This decision process is
unconscious, but it is not trivial. To characterize variability in the decision function, we adopt
the concepts of temperature and bias from  the field of reinforcement learning. The
bias captures the extent to which an individual generally favors conservative norms versus
innovations. Temperature characterizes the extent to which the individual chooses the optimal
production based on input received so far, versus exploring other options. A higher temperature means that individuals make
more random choices, an analogy to the behavior of gas molecules, which exhibit more random motion under higher
temperatures. In this application, we construe the optimal signal to be
the one that others are most likely to accept as the norm. A central innovation of
our model is a novel mathematical treatment of the decision function. In standard treatments of temperature, the
decision function is the SoftMax function, which reduces to the logistic in the case of binary choices.
The logistic is a sigmoidal function that maps inputs on the interval
 $(-\infty, \infty)$ to values on the interval $(0,1)$.  Varying the free temperature parameter $\tau$ causes its form to vary
between a step function as $\tau \rightarrow 0$ and a constant function as $\tau \rightarrow \infty$. 

We analyze the behavior of the logistic function in the context of the emergence of linguistic
norms, and identify serious problems with its  behavior and interpretability. Linguistic norms
resemble other application areas for reinforcement learning, because people learn them incrementally
from experiences over time. However, a critical difference is that the signals produced by each
individual provide the input for the neighbors, so that repeated signaling gives rise to positive
feedback loops in the community.  The logistic equation does not display sensible behavior under
such iteration. For any temperature $\tau > 0$,
it cannot characterize the situation in which an absolute agreement in the linguistic
community is stable. It is also unable to capture probability-matching behavior, in which
people produce linguistic variants with the same relative frequencies that they experience them.
Since probability-matching behavior
and stable agreement are two of the principle phenomena we need to characterize, we develop a new
sigmoidal decision function that incorporates the broad insights behind the logistic, while also
avoiding these problems. This is the {\it clog} (`{\it c}ognitive {\it log}istic') function. Like
the logistic, the $clog$ is a sigmoidal function that reduces to a step function as $\tau \rightarrow 0$. 
Unlike the logistic, it is designed to  map probabilities  onto probabilities. 
It converges to the line $y=x$, and not a constant function, as $\tau \rightarrow
\infty$. Variation in $\tau$ is usefully viewed as variation in the categoriality of the decision
process. As $\tau \rightarrow 0$, the output approaches two categories, $y = 1$ (or Boolean $TRUE$) and $y=0$ (or
Boolean $FALSE$). The variation in the input is completely regularized in the output. 
As $\tau \rightarrow \infty$, the categorization vanishes as all experienced probabilities
become equally available to the production system. The output faithfully reflects the input, without any regularization
at all. 

Finally, these ingredients come together in our treatment of the diffusion of  innovations as  {\it
informational cascades} \cite{bikhchandani}. Informational cascades occur in a social system when a local
change propagates through a large group as a result of people's repeated interactions and tendencies
to conform to each other. Initially developed in the area of opinion dynamics and product adoption
\cite{bikhchandani, watts2002, watts2007}, the concept is applicable to language dynamics under the assumption that linguistic
norms are collective opinions about how to express concepts \cite{fagyal}. Figure 1, the result of one
run of our model, illustrates a partial informational cascade in a social network in which people
vary in their biases. At the beginning of the run, the innovator (at the left edge) is the only
person in the network who has adopted the novel expression. Although the innovator is directly connected to only three other people,
the expression is adopted by many others in the network to varying degrees after a large number of communicative interactions have occurred.

\begin{figure}[!ht]
\includegraphics[width = 4.86in]{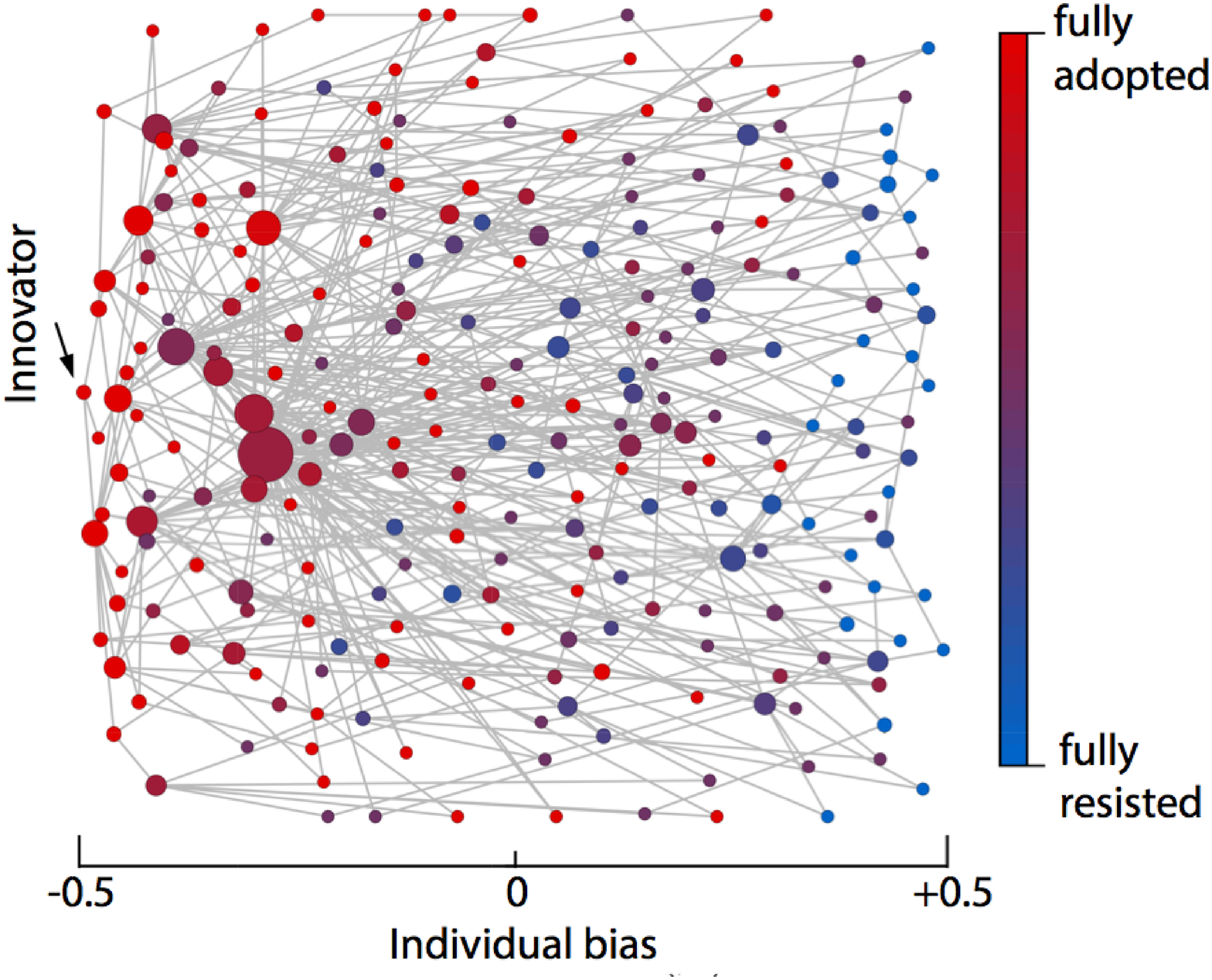}

\caption{{\bf A partial cascade of a novel form of expression originating from a low-degree innovator}. 
For each individual $i$,
the degree of preference $m$ for the novel opinion is represented as a probability from 
blue ($m_i = 0$) to red ($m_i = 1$). The network layout is governed by 
the distance from the innovator. Heterogeneous individual biases are assigned
on the basis of this distance as described in the text for the nearby scenario. The
decision rule is absolutely categorical (model
parameter $\tau = 0$).}
\label{Figure 1}
\end{figure}

To explore the behavior of the $clog$ in relation to likelihood of informational cascades in social networks, we report simulations
on networks of 256 nodes. This size corresponds to a village, social club, or
adolescent peer group that might  adopt a  new opinion, fashion, or
linguistic expression in a short time.  The social
network  is idealized as a scale-free network  generated by a preferential attachment rule \cite{albert}. 

We have already said that bias heterogenity is a key factor in the successful propagation of
linguistic innovations. Variation in individual biases is generated by sampling from a uniform
distribution. However, values are sampled in pairs to ensure the overall functional neutrality of
the innovative variant: each individual who is biased toward the innovation is balanced by a
different individual in the same run with an equal but opposite bias. In the field of opinion
dynamics, the adoption of an innovation is typically treated as absolute and irreversible (c.f. \cite{castellano, kleinberg}). 
Under these assumptions, bias heterogeneity has already proven to
be an important factor in the propagation of innovations \cite{delre}.
In contrast to these models, we treat individual changes as reversible, in the interests of empirical realism in the language domain. 
Reversibility does not affect which end states can be reached by the system, but it does 
impact the likelihoods of the different end
states, which is our primary interest. The previous observations about the importance of bias heterogeneity generalize to
this new model structure.

Our Monte Carlo simulation, implemented as an agent-based model, systematically explores the
interaction between categoriality, innovator degree, and the distribution of biases in the population.
We contrast two baseline scenarios, for which we expect rather stable norms, to three novel scenarios in which change is more likely.
In one baseline scenario, people simply reproduce the frequencies they encounter. This scenario is comparable to  
neutral evolution in evolutionary biology, and is already known to make incorrect predictions about rates of language change \cite{baxter}.
In a second baseline, people categorize the frequencies they encounter (to greater or lesser extents), but no one has any bias for
or against the change; the only heterogeneity in the population is in the number of social connections in the network. This setup is known to
be highly stable if the decision function is a step function \cite{delre,watts2002}, and we will extend this result to less categorical
decision functions. 
Of greatest interest are the three other scenarios, in which people in the network both categorize frequencies, and vary in their biases.
We designate these as the {\it random}, {\it hubs}, and {\it nearby} scenarios.
In the {\it random} scenario, bias values are distributed randomly in the community.
In the {\it hubs} scenario, innovation-favoring bias is preferentially allocated to high degree
nodes ({\it hubs}), and bias against the innovation is allocated to low degree nodes. In the {\it
nearby} scenario, innovation-favoring bias is preferentially allocated to the individuals
topologically nearest to the innovator (regardless of the innovator's degree), and bias against the
innovation is allocated to the individuals that are furthest from the innovation. 

Cascading is rare in the {\it
random} scenario, though this scenario does support some partial cascades beyond those that are observed in a neutral evolution model.
Two different regimes in which change is far more likely are
identified. In the hubs scenario, the probability of a cascade increases monotonically with
categoriality and innovator degree; total cascades become certain for sufficiently high
categoriality and innovator degree. This regime may be relevant for social phenomena in which a few highly connected individuals
are able to trigger sudden changes in collective opinion through their ability to broadcast their preferences to a large number of
people at once. However, it is less relevant to language change than the nearby scenario.
In the nearby scenario, cascades are most likely at a moderate level of
categoriality, and are most likely to be initiated by individuals with low to middle degrees of
connectivity. This  novel result captures the phenomenon of changes that are fast, arbitrary and
egalitarian. It represents a previously unsuspected regime for informational cascades. The discovery of
this regime is due to our model's ability to formalize the concept of moderate categoriality and explore the
interaction of categoriality with heterogeneity in the population. 

\section{Model definition and justification}

\subsection{Social network}

We consider $N$ individuals  in a community whose network structure is represented by a graph with adjacency matrix $A$. The matrix entry
$a_{ij} = 1$ if individual $i$ is a neighbor
of individual $j$ (e.g., individuals $i$ and $j$ are connected) and $0$ otherwise. 
As explained above, the links are interpreted to mean that the two individuals communicate with each other and are disposed to 
imitate each other; social connections that lack this force are omitted.
The number of  neighbors of individual $i$, $n_i = \sum_{j}a_{ij}$, defines the individual's {\it degree} in the network.
Individuals interact with their neighbors in the network, and these interactions are assumed to be bidirectional and uniformly weighted. Adjacency
matrices are generated using preferential attachment \cite{albert}, with an average degree of $4$. This represents an intermediate degree of network
connectivity that  falls within the range for global cascades to occur \cite{watts2002}. Results for average degree 3 are similar, and are not reported
here. 

\subsection{Social norms and signals}

Competing forms of expression are represented within the minds of individuals. With each communicative action, an
individual elects to use one of them. For 
example, to express the past tense of {\it slink}, a speaker of English has a choice between {\it slinked} and
{\it slunk}, but in each specific utterance, only one of these is used.  The choice
of signal is treated as a binary (Boolean) variable that assumes values of $0$ (FALSE) or $1$
(TRUE). In our simulations,  $1$ will always denote the innovation that may or may not cascade, and
$0$ will always denote the pre-existing consensus. This binary treatment of the available choices
entails little loss of generality. If a new expression, such as {\it snowpocalypse},
competes against a variety of pre-existing expressions, such as {\it blizzard}, {\it
large snowstorm} and {\it massive snowfall}, $0$ denotes the use of any of the earlier forms. 

We define the
observable signal from individual $i$ as a function of time as $s_i(t)$. The signal is communicated to all of the neighbors
in the network at the same time. This means that individuals with many neighbors (represented by nodes with high degrees)
are heard by many more people than individuals with few neighbors; however, by the egalitarian assumptions of the model,
these better-connected people also listen to more people.  The mental state $m_i(t)$ of an individual $i$ represents his or her private beliefs about norms of
language, as these beliefs develop over time. 
The model is initialized with $m_i(0) = 0$
for all individuals except a single innovator $v$, for whom $m_v(0) = 1$. The innovator could deviate
from the group consensus because his beliefs were formed before joining the network, or through a
mutation process whose details fall outside of the scope of this paper. As individuals acquire experience over time,
their mental states change because they are influenced by their impressions of their neighbors'
norms. We next describe  the learning rule that determines $m_i(t)$.  We return later to the characterization of
the relationship between $m_i(t)$ and the signal output at time $t+1$.

\subsection{Social assimilation}

In general, social learning is driven by general inclinations to assimilate one's opinions and behaviors to those of others in the community~\cite{watts2007,
castellano, bourhis, dijksterhuis}. 
Applying these ideas to language, individuals estimate the community norm from the relative frequency of each signal type in their aggregate
input. Taking $N_i$ to be the set of neighbors of individual $i$ (e.g. $\{j | a_{ij} = 1\}$)  and $n_i$ to be number of neighbors,
the input to that individual at each timepoint is thus the mean of the signals $s_j$ emanating from the members of $N_i$.
\begin{equation}
\label{eq:input}
input_i(t) = \frac{1}{n_i} \sum_{j | j \in N_i} s_j(t)
\end{equation}
However, individuals do not instantaneously copy the current input as their new mental state. People's beliefs generally reflect a combination of prior
experience and current input. Standard learning models capture this by using a weighted average of the prior mental state and the current input as
the update rule, with a {\it learning rate} parameter $0 \le \alpha \le 1$ \cite{lenaerts, bush}.
\begin{equation}
\label{eq:update}
m_i(t) = \alpha \cdot input_i(t) + (1-\alpha) \cdot m_i(t-1)
\end{equation}
The memory term
$(1-\alpha) \cdot m_i(t)$ holds individual $i$
back from slavishly following the neighbors.  $\alpha$ has no fixed
interpretation in relation to the time scale of learning, but rather reflects the temporal granularity of a
batch-processed approximation to social reality. We will report model runs
for $\alpha = 0.1$, selected to give insightful results using the
available computational resources. With this value of $\alpha$, individuals give $9$ times more weight to their existing knowledge than to
the immediate input from their neighbors.

\subsection{Unbiased decision rules}


The mental state $m_i$ and the learning rule Eq.~\ref{eq:update} capture the way in which the individual's beliefs about the ambient social norm evolve
as input arrives. Now we turn to the question of how individuals act on the basis of this knowledge, where action means
making a choice of which signals to express. We first consider the case in which individuals act without any kind of preference or bias.

\subsubsection{Criterion or threshold decision rules}

The criterion choice rule (also known as a threshold choice rule)
maximizes the expected utility in the case
of perfect knowledge \cite{sutton, friedman}. 
As a consequence of its mathematic definition, utility can assume values from $-\infty$ (an infinite loss) to $\infty$ (an infinite gain).
Considering a binary choice between mutually exclusive alternatives $a$ and $\neg a$, we define $Q(a)$ as the utility of $a$,
and  $ Q(\neg a)$ as the utility of $\neg a$. The criterion choice rule then takes the form:
\begin{equation}
\label{eq:steputility}
P(a) = %
\begin{cases}
0 & \text{if } Q(a) < Q(\neg a) \\
0.5 & \text{if } Q(a) = Q(\neg a) \\
1 & \text{if } Q(a) > Q(\neg a) \\
\end{cases}
\end{equation}
An individual should choose $a$ if it offers a net utility benefit over $\neg a$. 
The same equation can be also applied to perceptual decisions based on an
infinite scale of evidence; an individual should reach conclusion $a$ if the total evidence for $a$ is greater than the evidence for $\neg a$. 

For language, let's imagine that the individual wants to produce the signal that is most likely to be
accepted by any random member of the community. This means producing the signal that most other
people use, on the assumption that what people use is what they will accept. This goal is similar to the
goals of maximizing economic utility or reaching the best-supported conclusion,  except that the
input to the decision is not a utility or scale of evidence, but rather the $m_i$, the mental estimate of community norms.
The criterion choice rule can  be put into following form.

\begin{equation}
\label{eq:stepprob}
P(s_i = 1) = %
\begin{cases}
0 & \text{if } m_i < 0.5\\
0.5 & \text{if } m_i = 0.5 \\
1 & \text{if } m_i > 0.5 \\
\end{cases}
\end{equation}
The individual $i$ should choose $s = 1$ if, according to his beliefs, it is more likely than $s = 0$ to be the general norm. Eq.~\ref{eq:steputility} and Eq.~\ref{eq:stepprob} are
the same except for the range of the input variable. $Q(a)$ and $Q(\neg a)$  can take values anywhere
in the interval ($-\infty$, $\infty$), whereas
$m$ can only take values in the interval [$0$, $1$]. This difference in range has little importance if the decision function is a step function. However, for other
decision functions the difference in range has important consequences.

Eq.~\ref{eq:stepprob} is an absolutely categorical decision rule. Even if the input to the
individual has been variable, the individual treats one of the competing forms as  normative,
and the other input as deviant, invariably producing the normative form as the output. This is not
what people do. If linguistic norms vary amongst individuals in a group (as happens during periods
of language change), then many individual speakers also vacillate  in what form they choose to
express \cite{clark, kroch2001}. In experiments where some regularization of the input is found, it is never absolute \cite{culbertson, hudson2005, schumacher}.
To model real linguistic behavior, a less categorical decision rule is
needed.

\subsubsection{The SoftMax and logistic functions}

In economics, a threshold decision rule is only guaranteed to be optimal in the case of perfect knowledge. 
It may not be optimal in the case of imperfect knowledge, because the threshold must be
set on the basis of information already acquired, and it is possible that some information about the real optimum has not yet come to light. 
Speakers in a linguistic community also have imperfect knowledge of what other speakers do, because they only receive input from communications they are involved in. 
The statistical properties of communications that take place in more distant parts of their social networks are invisible to them,
as are the communications that will take place in the future. These 
observations lead naturally to the conjecture that the cognitive system processes the frequencies of linguistic inputs in a manner that
resembles reasoning under uncertainty in other areas of cognition.

The theory of reinforcement learning provides a general framework for understanding how learning
proceeds incrementally as more and more evidence becomes available from experience. The SoftMax
equation contains a free temperature parameter $\tau$ that captures the extent to which the
learner makes the optimal choice based on the evidence so far, or makes a random choice which may
yield more information in the future \cite{sutton}. If there are only two choices, the SoftMax reduces to
the logistic function Eq.~\ref{eq:logistic3}, which is a generalization of Eq.~\ref{eq:steputility}
and is expressed here in a form that emphasizes the parallelism.
\begin{equation}
\label{eq:logistic3}
P(a) = \frac{e^{Q(a)/\tau}}{e^{Q(a)/\tau} + e^{Q(\neg a)/\tau}}
\end{equation}

The parameter $\tau$ captures the categoriality of the decision.   As $\tau \rightarrow 0$, the learning system becomes 
completely categorical, rigidly
applying a threshold based on past evidence. This is the low temperature situation because the decision function can be viewed 
intuitively as a frozen reflex of the past.
As $\tau \rightarrow \infty$, the decision becomes random without regard to past evidence.

The logistic function is a sigmoidal function that is widely used in analyzing cognitive, social, and biological
processes that involve a competition  between two
states. Its many applications include not only decisions under uncertainty in economics, but also 
analysis of
the time course of linguistic 
changes \cite{chambers, tagliamonte}, the categorization of perceptual inputs \cite{falmagne, bogacz}, and
binary choices as a function of a continuous and unbounded scale of evidence \cite{friedman}. 

The crux of the problem is whether it makes sense to add
$\tau$ in just the same manner to a decision function that maps mental probabilities
onto signaling actions, as shown in Eq.~\ref{eq:logistic4}.

\begin{equation}
\label{eq:logistic4}
P(s = 1) = \frac{e^{m/\tau}}{e^{m/\tau} + e^{(1-m)/\tau}}
\end{equation}

The answer to this question rests on the behavior of Eq.~\ref{eq:logistic4} as it is iterated through repeated social interactions. 
Figure 2 graphs Eq.~\ref{eq:logistic4} as the temperature $\tau$ is varied over its full range. To understand
the behavior of the function,  it will be convenient to have
an equation for the angle $\phi$ at the inflection point. 

\begin{equation}
\phi = \arctan\frac{1}{2 \cdot \tau}
\end{equation}

Varying $\tau$ is accomplished by varying
$\phi$ from $90^\circ$ to $0^\circ$. $\phi = 90^\circ$ in the limit as  $\tau \rightarrow 0$ 
and $\phi =  0^\circ$ in the limit as $\tau \rightarrow \infty$.
The points of most interest are the fixed points,
defined as the points for which the output and input are the same (eg. the points where
Eq.~\ref{eq:logistic4} crosses the identity function $P(s_i=1) = m_i$). The stable fixed points 
act as  attractors; that is, slight deviations in the vicinity of a stable fixed point push the result back onto the same
point. The unstable fixed points act as repellers; slight deviations send the result away
from the fixed point. Thus, the stable fixed points represent predictions about the 
possible long-term states for the system. 

\begin{figure}[!ht]
\includegraphics[width = 6.5in]{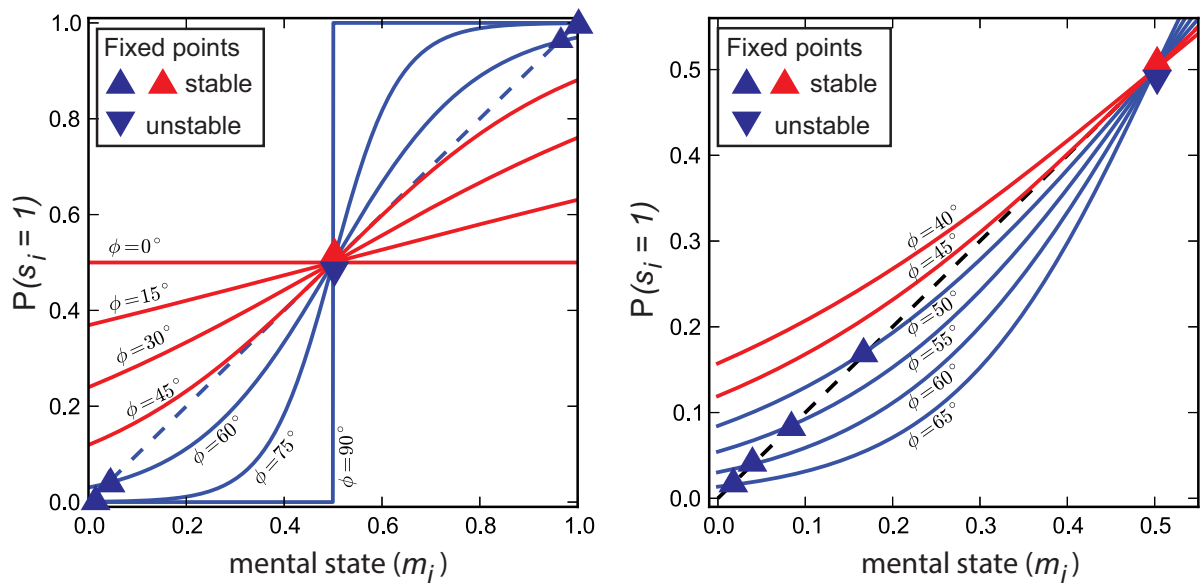}
\caption{{\bf Behavior of the {\it logistic} function as $\phi$ is varied.} $\phi$ determines the location and nature of the fixed points.
{\bf A}: Except for $\phi = 90^\circ$, the function lacks fixed points at $(0,0)$ and $(1,1)$. For $\phi \leq 45^\circ$, the unstable fixed point at
$(0.5,0.5)$ becomes a stable fixed point. {\bf B}: Detailed
view of behavior in the lower-left quadrant. $\phi$ varied in $5^\circ$ steps. For $\phi > 45^\circ$, there is an unstable fixed
point at $(0.5,0.5)$. For $\phi \leq 45^\circ$, the fixed point at $(0.5,0.5)$ is stable.}
\label{Figure 2}
\end{figure}

The location and stability of the fixed points for Eq.~\ref{eq:logistic4} vary as $\phi$ is varied. 
For $45^\circ < \phi < 90^\circ$, there is
an unstable fixed point at $(0.5, 0.5)$, which means that for moderate 
temperatures, free variation between two competing forms is unstable towards a system in
which one variant is preferred. This behavior is somewhat realistic, but not
entirely so, because the location of the stable fixed points is problematic. 
No value of $\phi$ other than $90^\circ$
has a stable fixed point at $(0,0)$ or $(1,1)$. For $45^\circ < \phi < 90^\circ$, there are indeed two stable fixed points.
However, one is always located at $0 < m_i < 0.5$ and the other is always located at $0.5 < m_i < 1$.

This situation has the unfortunate consequence that inputs of $m_i = 0$ and $m_i = 1$ yield outputs
that are shifted towards the center. It means that all binary competitions in language continue to
the end of time; no form or expression can ever truly go extinct. To make a comparison to norms of attire, some 600
years ago, garments called {\it chausses} were in competition with garments called {\it breeches};
{\it breeches} eventually won out. If neither I nor anyone in my community has any knowledge or
experience of {\it chausses}, then we have all adopted the innovative
form and a model of our speech community has $m_i = 1$ for all individuals $i$.
A model using a logistic choice function would nonetheless predict that the archaic form {\it chausses}  has a 
predictable and persistent tendency to reappear as a competitor to {\it breeches}, as we continue to interact with
each other in the future.  In fact in language, just as in biology,
something that is extinct is gone forever. New competitions arise, but these are due to fresh
innovations. The invention of {\it trousers} is what created later competition for {\it breeches},
and not the persistent return of {\it chausses} from the graveyard of fashion history.

A further problem is revealed for $0^\circ \leq \phi \leq 45^\circ$. In this case, the fixed point at $(0.5, 0.5)$ becomes stable. At high
temperatures, the system is predicted to converge towards free variation. 
For language systems, this asymptotic behavior is highly questionable. This would mean that if people weakly categorize the input they
receive about two competing words or constructions, their output persistently reverts towards completely free variation. This would
behavior would count as a tendency towards true synonymy (in which two forms completely share their denotation and their range of use).
But, in the psycholinguistic literature, avoidance of true synonymy is documented in people of all ages, and it is argued to play a key role
in the emergence of meaning categories in language \cite{clarkE, davis2009, baronchelli}.
The formalization of the concept of temperature
in the context of language and collective opinions needs to be reconsidered.

Note finally that the logistic never falls on the line $y=x$. For $\phi = 45^\circ$, it tracks $y=x$ in the middle of the range, but curves away from
this line for more extreme values. This means that it cannot be used to model probability-matching behavior, 
which is widely assumed to be relevant for language \cite{labov1994, griffiths, reali, nam}.

\subsection{Introducing the $clog$ function: categoriality and bias}

The problems with the applicability of Eq~\ref{eq:logistic4} arise from the fact that it converges to $P(s = 1) = 0.5$ as $\tau
\rightarrow \infty$. 
Although Eq.~\ref{eq:steputility} and
Eq.~\ref{eq:stepprob} have similar behavior as formulations of a criterion (or expectation maximization)
decision rule, this similarity disappears as soon as nonzero temperatures are introduced. This
problem is solved if we assume that the sigmoidal function should instead approach $P(s_i) = m_i$ as
$\tau \rightarrow \infty$. We now develop the mathematical apparatus to explore this assumption.

We define a new relative of the logistic, the {\it clog} (`{\it c}ognitive {\it
log}istic') function that converges to the identity function as $\tau \rightarrow \infty$: 

\begin{equation} 
\label{eq:mlog1} 
clog_{\tau}(m) = \frac{m \cdot e^{m/\tau}}{m \cdot e^{m/\tau} + (1-m) \cdot e^{(1-m)/\tau}} 
\end{equation} 

Just as for the $logistic$, it will be convenient to characterize
the degree of nonlinearity in the $clog$ function in terms of the slope $\phi$ at the inflection point, which we 
will denote as $m^\ast$. It is:
\begin{equation}
\label{eq:phiclog}
\phi = \arctan{(1+ \frac{1}{2 \cdot \tau})}
\end{equation}
 
In the limit of $\phi \rightarrow 45^\circ$ (i.e. $\tau \rightarrow \infty$), $clog$ reduces to the identity function; the
production probabilities are exactly the same as the current mental state.
As $\phi \rightarrow 90^\circ$ (i.e. $\tau \rightarrow 0$), the function
approaches a step function with the discontinuity at $m^\ast$.
For all $\tau < \infty$ ($\phi > 45^\circ$),  Eq.~\ref{eq:mlog1} has attracting fixed points at $(0,0)$ and $(1,1)$. This
means that uncertain opinions tend to become polarized in the course of repeated interactions towards more
absolute opinions. The stability of these fixed points provides for the ultimate extinction of losing competitors.
The inflection point $m^\ast$ in between  $m = 0$ and $m = 1$ (located at  $m = 
0.5$) is a repelling fixed point. Because $m^\ast$ is always unstable, there is never a regime in which the
system is attracted towards free variation. 

The limiting case represented by $\phi = 45^\circ$ is studied in the literature under the name of a {\it probability-matching} rule,
{\it a relative goodness} rule or  {\it Luce's Choice
Rule} \cite{friedman}.

Thus far, we have considered only unbiased decisions, which rest entirely on the balance of utility, evidence, or likelihood of the 
two alternatives. In practice, biased decision-making processes are frequently encountered. Such biases are normally accommodated in 
Eq.~\ref{eq:logistic3} for the $logistic$ by introducing a second free parameter $\beta$.

\begin{equation}
\label{eq:logisticbiasQ}
P(a) = \frac{e^{{(Q(a)-\beta)}/\tau}} {e^{{(Q(a)-\beta)}/\tau} + e^{{(Q(\neg a) +\beta)}/\tau}}
\end{equation}
Varying $\beta$ has the effect of shifting the entire function. If $\beta < 0$, the individual is more likely to pick
the innovative variant than would be expected from its utility. If $\beta > 0$, he is less likely to do so.

Figure 3 illustrates the results of attempting to recast Eq.~\ref{eq:logisticbiasQ} as a function from mental probabilities to signaling actions,
as shown in Eq.~\ref{eq:logisticbiasm}.

\begin{equation}
\label{eq:logisticbiasm}
P(s = 1) = \frac{e^{{(m-\beta)}/\tau}}{e^{{(m-\beta)}/\tau} + e^{{(1 - m +\beta)}/\tau}}
\end{equation}

\begin{figure}[H]
\includegraphics [width = 4.86in]{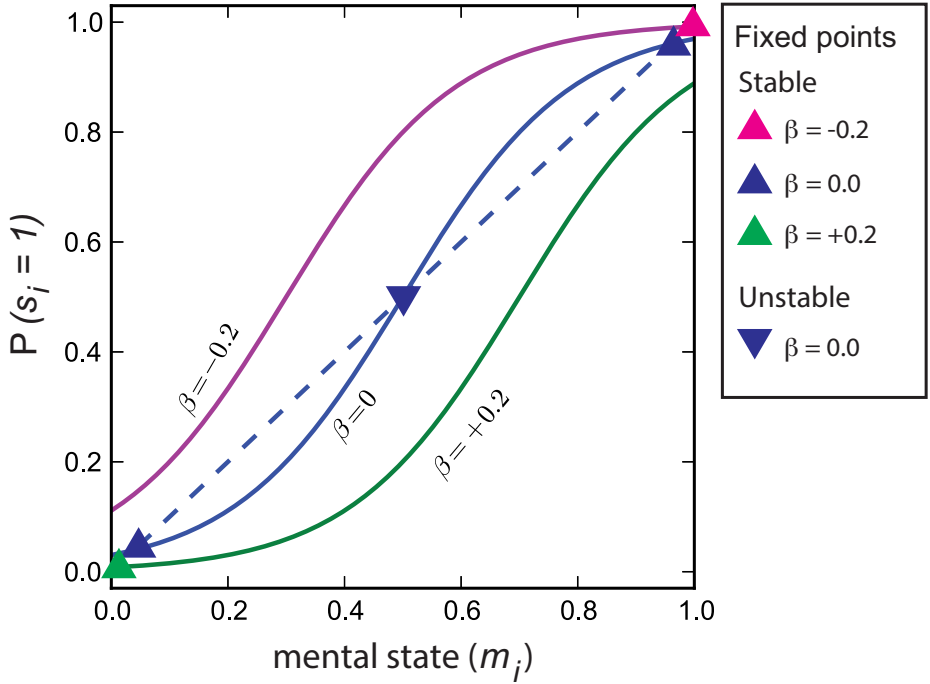}
\caption{ {\bf Behavior of the {\it logistic} function for $\phi = 60^\circ$ as $\beta$ is varied.} The unstable fixed point at $(0.5, 0.5)$ is lost
for small biases $\beta = 0.2$ and $\beta = -0.2$. Each of these $\beta$ values results in a system with a single stable fixed point.}
\label{Figure 3}
\end{figure}

The shift in the function shifts the location of the inflection point occurring in the
unbiased case at $m_i = 0.5$.  The number and stability of the fixed points is greatly affected. A small positive bias gives
rise to a single stable fixed point representing a high level of acceptance of the innovation, and a small negative bias yields a stable
fixed point that represents a low level of acceptance of the innovation. 

In contrast to the $logistic$, the $clog$ can incorporate a bias parameter without restructuring the fixed points. 
Including a bias parameter $\beta$ into the $clog$ function yields the following fully elaborated form.

\begin{equation} 
\label{eq:mlog} 
clog_{\tau, \beta}(m) = \frac{m \cdot e^{(m-\beta)/\tau}}{m \cdot e^{
(m-\beta)/\tau} + (1-m) \cdot e^{(1 -m + \beta)/\tau}} 
\end{equation} 

For the $clog$, $\beta$ controls the location of the unstable fixed point $m^\ast$, which is always located at $m_i = 0.5 + \beta$. 
If $\beta_i < 0$, $m_i^\ast < 0.5$, and individual $i$ has a greater propensity to produce $1$ than the input from the
neighbors would dictate. Similarly, if $\beta_i > 0$, $m_i^\ast > 0.5$, and the individual has a disproportionate tendency to produce $0$. 
The stable fixed points are unaffected by the value of $\beta$. The introduction of a function that smoothly interpolates between 
a criterion choice rule and a probability-matching choice rule, under under any choice of bias, 
while always having fixed points at $(0,0)$ and $(1,1)$, is a novel contribution of this paper.

Heterogeneity amongst community members in $\beta$ is the primary type of heterogeneity that we explore. The different scenarios for
informational cascades depend on positive and negative values of $\beta$ being distributed in different ways across the individuals
in the network. 

Figure 4
illustrates the behavior of $clog$ as $\phi$ and $\beta$ are varied. In this figure, note particularly that varying $\beta$ has no 
effect on the fixed points at $(0,0)$ and $(1,1)$ for $\phi > 45^\circ$. This entails that the $clog$ (unlike the $logistic$)
cannot represent a uniform distribution on the closed interval $[0,1]$. 
Note also that the effects of $\beta$ are neutralized if
$\phi = 45^\circ$; the concept of bias is only meaningful if the choice rule is at least somewhat categorical. 

\begin{figure}[!ht]
\includegraphics[width = 4.86in]{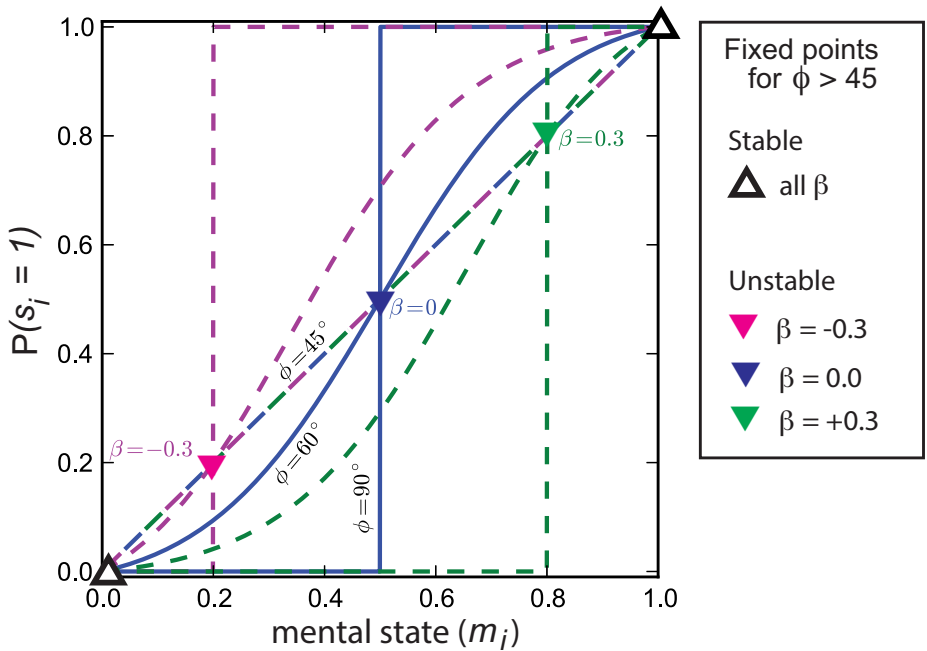}
\caption{{\bf Behavior of the {\it clog} function as $\beta$ and $\phi$ are
varied.} Blue: No bias. Magenta: Bias against the pre-existing consensus, equivalent to bias 
toward the innovation. Green: Bias toward
the pre-existing consensus, equivalent to bias against the innovation.
When $\phi = 45^\circ$, the {\it clog} reduces to the dashed line $P(s_i=1) = m_i$ for
all values of $\beta$.}
\label{Figure 4}
\end{figure}

\subsection{Implementation}

The model is implemented in the NetLogo agent-based modeling environment \cite{uri}. It 
is initialized at time $t = 0$ with $m_i(0) = 0$ for all individuals $i$ in $A$,
except for a single innovator $v$, for whom $m_v(0) = 1$. 
During each model cycle $t+1$, every individual first randomly produces
an expression of opinion, $0$ or $1$, with probability: 
\begin{equation}
\label{eq:production}
P(s_i(t) = 1) = clog_{\tau, \beta}(m_i(t))
\end{equation}
Then they update their
mental state from $m_i(t)$ to $m_i(t+1)$ on the basis of the set of signals emitted by
their neighbors $N_i$ (e.g. $\{s_j(t) | j \in N_i\}$), using Eq.~\ref{eq:update}.
Each model run is terminated when the population reaches a consensus (defined
as $m_i < 10^{-8}$ or $m_i > 1 - 10^{-8}$ for all $i$), or at the 10,000th iteration
if no convergence has occurred.  

The probabilities of cascades as a function of innovator degree are estimated by a Monte Carlo
method, in which innovator degrees from 2 to 55 are sampled equally in the form of 500 runs for each degree.
Innovators with degrees of $1$ and $> 55$ are not included because the preferential attachment
rule for network construction does not generate nodes with such degrees frequently enough.  
A fresh network is generated for each run, and all individuals share a value of $\phi$. $\phi$ was 
smoothly varied from $\phi = 45^\circ$ (linear) to $\phi = 90^\circ$ (categorical) by increments of $1^\circ$.
In both baseline scenarios, all individuals are unbiased. (In the first baseline, this is because they
follow a probability-matching decision rule.
In the second baseline, the decision rule is more categorical and $\beta_i = 0$). In all other scenarios,
heterogenous values of $\beta_i$ are assigned, as described in the text.

\subsection{Summarizing outcomes}

In our Monte Carlo simulations, we vary the categoriality,
the degree of the innovator and the way that bias values are distributed across the
community. Each model configuration
yields a distribution of cascade sizes; the 
size of a cascade is distinct from the likelihood of a cascade. In order to summarize the results, however, we 
conflate these two factors. We 
use the following classifications of the average mental state $\bar{m}$ of
all the individuals in the network when the model run ends at time $t_{final}$.
\begin{itemize}
\item {\it Survival}: The innovation avoids extinction, defined as $\bar{m}(t_{final}) > 10^{-4}$.
\item {\it Dominance}: The innovation becomes dominant by surpassing the established variant, i.e. $\bar{m}(t_{final}) \geq 0.5$.
\item {\it Completion}: The innovation drives the
previous consensus opinion to extinction i.e. $\bar{m}(t_{final}) \geq 1 - 10^{-4}$.
\end{itemize}

\section{Results in different scenarios}
We first present two
baseline cases in which all individuals are unbiased. 
Then, we compare three
scenarios in which individuals have heterogeneous biases.

\subsection{Baseline scenarios}
If $\phi = 45^\circ$, each individual expresses each opinion in direct proportion to their experience. The model instantiates the case of {\it
neutral evolution} (also known as {\it random drift} \cite{baxter}). For a fully continuous system, the population would converge to the average of
the initial states of the individuals; however, because emitting a signal is a discrete probabilistic event, the mean  state instead exhibits a
random walk on the frequencies of the alternative signals $0$ and $1$ \cite{baxter}. The probability of a complete cascade is low, and is directly
proportional to the innovator's degree, as shown in Figure 5. 

\begin{figure}[!ht]
\includegraphics[width = 3.27in]{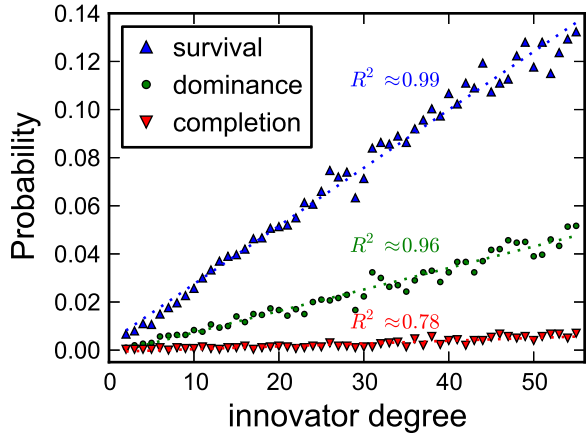}
\caption{{\bf Relationship of innovator degree to cascade probability in
the neutral evolution model.} For all three summary cases, the relationship
is linear. The probability of a complete cascade is extremely small.}
\label{Figure 5}
\end{figure}

If $\phi > 45^\circ$ (e.g individuals have at least
some tendency toward categorical behavior), and all individuals are unbiased ($\beta_i = 0$ for all $i$), 
then the innovation always becomes extinct; no cascades of any size are found.
This very strong result on the impossibility of change without
bias heterogeneity occurs because memory is included in the model of learning. Eq.~\ref{eq:update} entails that a single signal $s_i = 1$ from an
innovator $i$ can never cause the mental state of a neighbor $j$ to increase from its initial value  $m_j = 0$ to a value greater than the repelling
fixed point at $m_j^\ast = 0.5$. The innovator cannot convert anyone else, and is eventually reconverted to the original consensus.

\subsection{Scenarios with heterogeneous biases}
To explore the role of heterogeneous biases while maintaining overall
functional neutrality, bias values are generated through random sampling on a uniform distribution over the interval $[0,0.5]$.
Neutrality is enforced by taking both the $+\beta$ and $-\beta$ for every $|\beta|$ that
is selected. The
resulting set of values is distributed over the individuals in the social
network according to one of three different methods. In the {\it hubs}
scenario, the bias against the pre-existing consensus is preferentially allocated as a
function of degree centrality. 
The highest degree individuals are most disposed
to change, and the lowest degree individuals are least disposed to change. In
the {\it nearby} scenario, the bias against the pre-existing consensus is
allocated as a function of the distance from the innovator; individuals nearest
to the innovator are the most disposed to change, and those farthest from the
innovator are least disposed to change. In the {\it random} scenario,
bias values are randomly distributed.  

The hubs and nearby scenarios idealize skewed patterns in the distribution of
biases in the network. To motivate these
idealizations, let us consider what the bias $\beta$ means.
The case of negative bias generalizes the concept
of early adopters, defined in \cite{watts2002, watts2007} as
individuals who will adopt an innovation after encountering a single example of it. Negative bias means that an individual is inclined to abandon 
the old norm, even if most of the neighbors do
not. Positive bias is a generalization of the concept of late adopters. It
means that an individual is inclined to continue using the old expression, even
if the neighbors use its more innovative competitor. 

Such individual biases could arise
as a reflex of personality traits such as levels of adventurousness and nonconformity, which
are factors in the standard theory of personality \cite{john} and are partially
innate in people and other animals \cite{wilson}. The differences could also
arise from knowledge or habits that 
provide a latent potential or impediment to the adoption of
the innovation. As a simple example, the use of text messaging provides a platform for
the use of slang acronyms like {\it lol} (laughing out loud) and {\it btw} (by
the way), and  it would not be surprising if heavy users of text messaging were
among the first to adopt these expressions upon encountering them. 
In modern linguistic theories, the adoption of one construction may also provide a foundation for further change
through its indirect impact on the encoding system \cite{kroch2001}.
It is possible to develop a connection between
such individual biases and the concept of utility. We already provided an interpretation of $\tau$ in relation to
the global value of addressing any member of the community in a cooperative manner, 
using a form they are likely to accept. $\beta$ encapsulates
any sort of subjective personal utility that is independent from the utility to the community as a whole of having shared norms.

The distribution of biases in the hubs scenario thus can be understood as one in
which individuals with many social connections
share any type of knowledge, practice, or experience that makes it easy for a specific
linguistic innovation to take root. The nearby scenario is even more plausible, because
people with shared traits tend to form social connections to each
other \cite{wasserman}. Field studies of Belfast English indicate that feelings of group solidarity affect people's
choice of modes of expression \cite{milroy}. 
The rate of adoption of linguistic innovations
peaks in adolescence \cite{tagliamonte}. In choices of language, music, and attire, adolescents tend to be most 
influenced by the central members of their immediate peer group (rather than by the highest status people in their whole
world of experience) \cite{labov1972, eckert, mendoza}. A group of young people who share rebellious  attitudes and a love of
novelty is indeed the very prototype of a group with innovative language. 

Results for the three scenarios are shown in Figure 6. 

\begin{figure}[H]

\includegraphics[width = 6.5in]{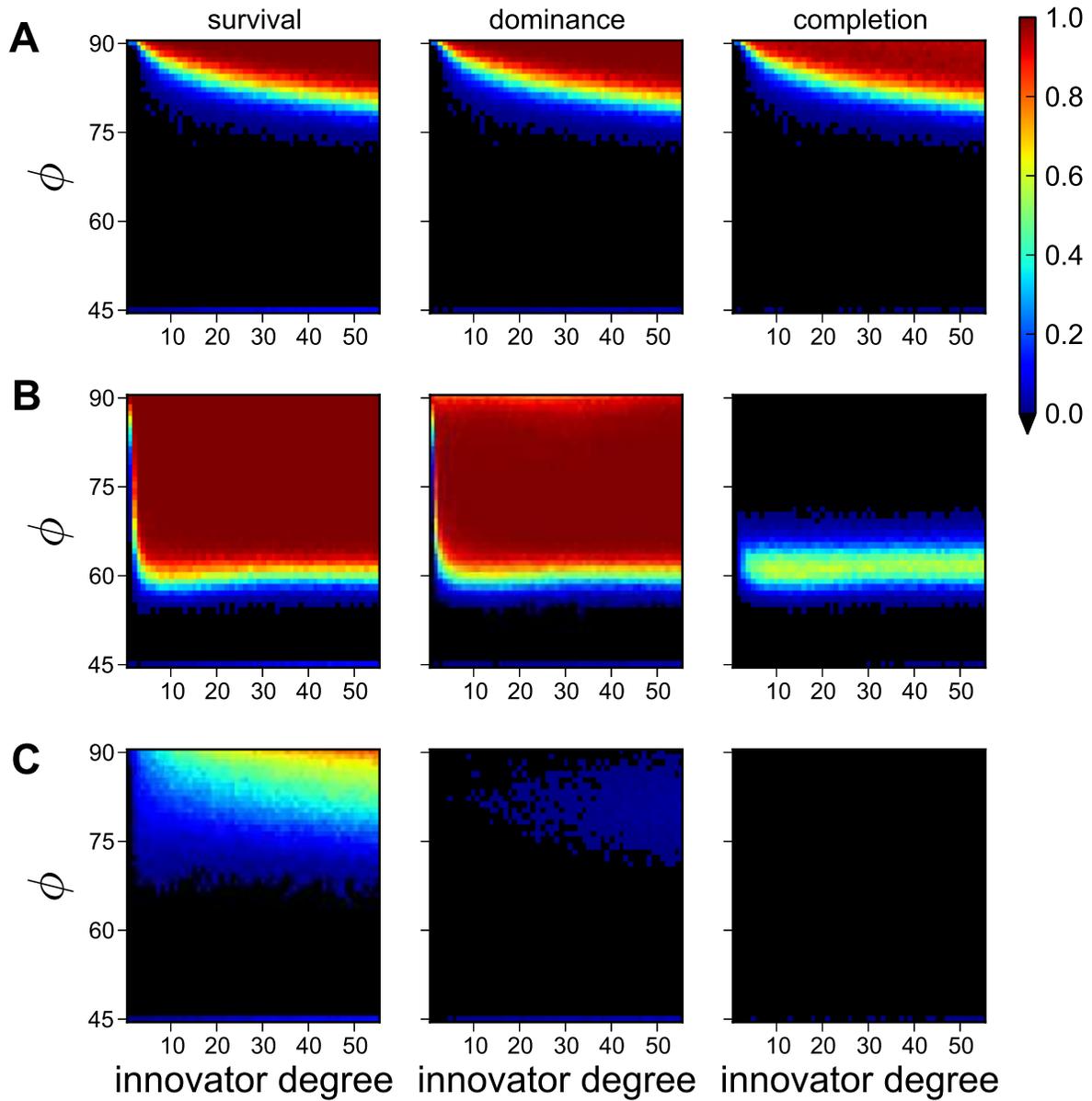}
\caption{\bf Interaction of bias distribution with innovator degree and
categoriality  in determining the likelihood of cascades.} 
Non-zero likelihoods of a cascade in each of the three categories survival,
dominance, completion are color-coded as shown. Black indicates the absence of
any cascades in the simulation results. {\bf A}: Bias values are assigned to individuals as a function of
degree centrality under the hubs scenario. {\bf B}: Bias values are assigned to individuals as a function
of proximity to the innovator under the nearby scenario. {\bf C}: Bias values are assigned
randomly to individuals. 
\label{Figure 6}
\end{figure}

\subsubsection{Results for the hubs scenario}

In the hubs scenario (Figure 6A), cascading is
frequent and consistent for high-degree innovators and high $\phi$ values.  
As the innovator degree decreases, the likelihood of success also decreases.
Furthermore, cascading is basically an \textit{all-or-nothing} phenomenon; if a
cascade avoids extinction, it is very likely to go all the way to completion.
Intuitively, if an innovation survives the initial stage in this scenario, it
has done so by spreading to a backbone of high-degree adopters. These in turn
can dominate the overall signal to the rest of the network, because each of
their expressions of opinion is sent to many neighbors. Interestingly, when the
input-output relation is not strongly categorical (i.e., $\phi \lessapprox
75^\circ$), assigning the favorable bias values to the hubs is \textit{not} effective
for causing cascades (regardless of innovator degree). As seen in Figure 6A, a
phase transition occurs in the high-$\phi$ regime ($\phi> 75^\circ$). As
categoriality increases, the outcome passes from invariable cascade extinction,
through degree-sensitive all-or-nothing cascades, to complete cascades
(apart from very low-degree innovators).

Just visible in blue at the bottom of the graph is a sporadic set of cascades arising from neutral evolution ($\phi = 45^\circ$), 
corresponding to those displayed in Figure 5.

\subsubsection{Results for the nearby scenario}

We identify three regimes in the nearby scenario (Figure 6B).  
For $45^\circ \le \phi
\lessapprox 55^\circ$, no cascades occur, apart from the few previously discussed for
the $\phi = 45^\circ$ neutral evolution model. For $55^\circ \lessapprox \phi \lessapprox
65^\circ$, cascades are more frequent, and often go to
completion. For $65^\circ \lessapprox \phi \leq 90^\circ$, cascades become
increasingly likely, but do not go to completion even if they succeed in
dominating the network. In particular, in the fully categorical limit, where
$\phi = 90^\circ$, the cascade  proceeds only a little further than
halfway: $\bar{m}(t_{final}) = 0.56$ (with the standard deviation $\sigma =
0.069$). The network already displayed in Figure 1 is a final configuration (at
the 10,000th time step) for this partial cascade situation. It shows how the
novel opinion diffuses through some (but not all) of the network, leaving an
entrenched opposition elsewhere. This means that for innovations occurring
in a social network with a nearby bias distribution, a highly categorical decision rule tends to 
split the social network into subgroups with different norms.

In-depth examination of the case $\phi=60^\circ$ (Figure 7) reveals another striking point. The likelihood of success in triggering a cascade peaks in the
vicinity of innovator degrees $7 \leq n_i \leq 15$ for this level of categoriality. The individuals 
with degrees in this range are more likely to succeed than very
low-degree individuals, an effect also visible for other values of $\phi$ at the left edge of the panels in Figure 6B. Surprisingly, they are also
somewhat more likely to succeed than high-degree individuals. With this combination of categoriality and bias, it is clearly not appropriate to
describe the best-connected nodes as \textit{influentials}; their influence is actually constrained by the variability of their many neighbors.
Taking the network's degree distribution into account, we can also calculate the probability of the innovator degree, given that a
cascade occurred. As Figure 7 shows, this probability  peaks at innovator degree $n_i = 4$,  which is by construction the average degree of the network. Thus 
in this scenario, a successful innovation is most likely to have originated from Joe or Jane Average.

\begin{figure}[!ht]
\includegraphics[width = 3.27in]{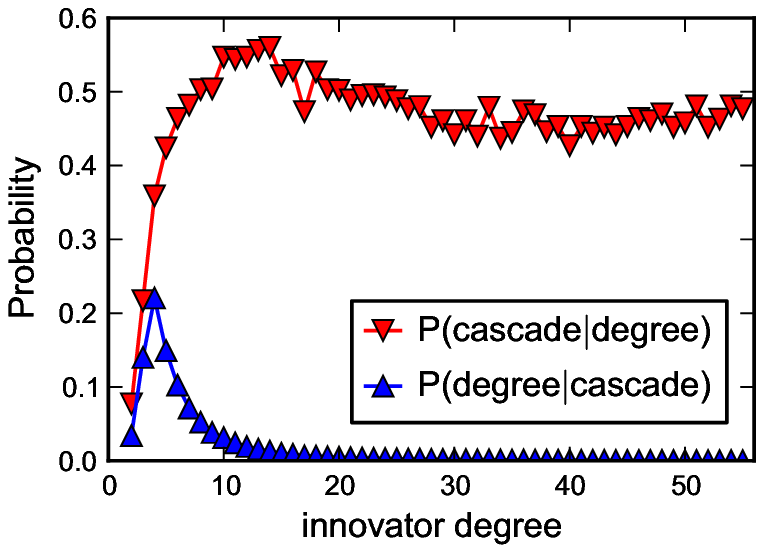}
\caption{{\bf Relationship of innovator degree to complete cascades for the nearby
scenario with} ${\boldsymbol{\phi}  {\boldsymbol =} {\boldsymbol 6}{\boldsymbol0}^{\boldsymbol \circ}}$. Red: The probability of a complete cascade, given the degree of the innovator.
Blue: The probability of the innovator degree, given that a complete cascade occurred. Each data point is
estimated from the 500 runs per degree in the Monte Carlos simulation.
Results are highly similar for cascade survival and dominance.}
\label{Figure 7}
\end{figure}

\subsubsection{Results for the random scenario}
The hubs and nearby scenarios are highly idealized, and yield exorbitantly high success rates for innovation. In reality,  the bias distribution is
probably less stringently tied to network position, and very many innovations fail. A perspective on this reality is provided by results on the {\it
random} scenario in Figure 6C. Survival of an innovation is fairly common when $\phi \gtrapprox 70^\circ$. The rate of survival is notably higher than in
the neutral evolution model as shown in Figure 5, also visible in this figure panel as a blue line at $\phi = 45^\circ$. Dominating cascades are more rare.
In addition to the few dominating cascades found for $\phi = 45^\circ$,
a small cluster of dominating cascades is found for degree $n_i \gtrapprox 10$ and $\phi \gtrapprox 80^\circ$. Thus,
moderate categoriality in combination with heterogeneity permits a moderate to high degree innovator to have some chance of seeing an innovation
widely adopted. This likelihood becomes attenuated as $\phi$ approaches $90^\circ$. Qualitatively, the pattern suggests a mixture of cases skewed towards
the nearby scenario and cases skewed towards the hubs scenario. Complete cascades were not observed at all above $\phi = 45^\circ$.  Of course, a complete
cascade could occur if the random bias distribution happened to distribute the bias in exactly the same way as in one of the more favorable scenarios in
Figure 6AB. The fact that this outcome is too rare to occur at all in our simulations suggests that bias heterogeneity alone cannot fully explain how
neutral innovations may come to be adopted by an entire community. Instead, systematic patterns are needed
in the distribution of decision biases across individuals.

\section*{Discussion}

We have presented a framework for exploring the interaction of categoriality, bias, and innovator
degree in permitting arbitrary linguistic innovations to succeed in a social network. As anticipated
from prior related models \cite{baxter, delre, watts2007}, model configurations in which individuals
are homogeneous and unbiased exhibit negligible success rates for innovations. We identified two
regimes involving heterogeneous biases that are conducive to change. Under the hubs scenario, a
highly categorical decision rule supports a mechanism for the widespread adoption of innovations initiated by highly connected
people.  Under the nearby scenario, a moderately categorical decision rule leads to a
mechanism for the grassroots changes to succeeed. This second regime, which was previously unsuspected in
research on informational cascades, is highly relevant to linguistic change
because linguistic changes typically originate with ordinary people.

Previous models of language dynamics have explored the ways that functional factors can have major
impacts on the linguistic structures through their iterated effects in language acquisition,
perception, and production \cite{griffiths, croft}. Here we have shown that such functional factors
are not necessary for global changes to occur. Our results connect the understanding of language
dynamics with classic observations about the arbitrariness of many words and expressions in human languages\cite{newmeyer, saussure}.

Statistical variability in language is pervasive and often persists through language learning from one generation to
the next \cite{labov1994}. Many researchers have assumed that the cognitive system for language uses probability-matching decision
rules \cite{nam, reali}. However, tendencies
toward regularization of observed frequencies have been found in the area of phonotactics \cite{frischthesis, hudson}, and
in some experiments involving learning of artificial languages \cite{culbertson, hudson2005}.
It has been argued that people discount ambiguous evidence in the
acquisition of morphology and syntax \cite{kirby, pearl}. 
Ambiguity and vacillation may be cognitively costly
for linguistic processing. Categorical processing is argued to support fast, accurate encoding and
decoding, as well as the ability to create new words and sentences by recombining basic
elements \cite{deboer}. 

Our results reconcile these disparate threads in the literature by developing a formal apparatus for analyzing
regularization as the categorization of experienced frequencies. We then
suggest that the cognitive processing of frequencies in language is moderately categorical. 
The regime in which we observe grassroots changes has $\phi \approx 60^\circ$. As shown in Figure 4, this is a level of
categoriality that deviates only moderately from a probability-matching decision rule. Although this moderate nonlinearity
is sufficient to completely change the prospects for an innovation to succeed, it is probably too weak to be detected in a typical
psycholinguistics experiment. Very large scale data collection over the entire range of input probabilities would 
also be needed to assess the heterogeneity in individual biases that also plays a key role in our model. In short, the statistical power
needed to rigorously compare a probability-matching (or neutral evolution) model 
with the regime for complete cascades in the nearby scenario is much higher than has been available in many classic experimental studies.

Why isn't language processing strongly categorical? A probability-matching decision rule can be
optimal at the group level when coordinated actions are needed to divide up
resources \cite{labov1994}. Language presents a related case, because coordinated actions are
needed for a group to maintain a mutually intelligible language, which benefits the group as a whole
by permitting knowledge to be transmitted from any person to any other. In the nearby scenario, a
completely categorical decision rule jeopardizes mutual intelligibility. For $65^\circ \lessapprox
\phi \leq 90^\circ$, arbitrary innovations dominate the system but do not go to completion. This
outcome means that the social group has split into subgroups with distinct norms. In contrast, the
regime $55^\circ \lessapprox \phi \lessapprox 65^\circ$ yields a greater likelihood that mutual
intelligibility will be maintained through periods of change. Assuming that the distribution of
biases in the nearby scenario is empirically well-founded, language processing can be understood as
striking a balance between fast, accurate processing and the maintenance of a shared linguistic code.

In research on opinion dynamics, an early and prominent hypothesis held that a small number of
individuals with high degree in a social network had disproportionate importance in triggering
global cascades. That is, hubs in the network were thought to have special importance as so-called
{\it influentials}~\cite{watts2007}. Some previous models of language change 
achieve realistic cascade rates by assuming that people weight
positively the input from highly-connected people, who are also assumed to be high-status \cite{nettle, fagyal}. In the extreme,
these assumptions lead to the conjecture that broadcast media should be a major factor in language
change, since the media transmit the  linguistic patterns of political leaders and celebrities to such large numbers of
people. This conjecture is not well supported. For example, establishing any effect whatsoever of
television on local dialects has been an elusive goal, and only recently has such an effect been
found. In a detailed field study of Glaswegian English, some 
TV watchers were found to be influenced by the East End dialect of London, but only if they 
had personal ties to the East End or felt strong psychological
engagement with characters in a popular TV soap drama that is set there \cite{stuart}.

In our model of egalitarian changes (the nearby scenario), informational cascades occur without any
positive weighting of input from highly connected individuals. These individuals do play an
important role, in that successful innovations succeed in part by being adopted by one or more
highly connected individuals, who then broadcast them to many people. On the average, however, the
highest degree individuals are somewhat conservative compared to less connected individuals, because
their linguistic patterns are stabilized by the large number of incoming signals. Individuals who
are far from the innovator, and biased against the change, eventually adopt it with some probability
simply because a core of positively biased individuals near the innovator allows the change to gain
traction and in some cases eventually dominate the signal to the whole community. Some previous work
has argued that the concept of {\it covert prestige} (understood as an implicit value system that
countervails overt norms) is necessary to explain why high status people often adopt stigmatized
expressions originating in lower status circles \cite{trudgill}. However efforts to substantiate
this notion in terms of perceived lower-class attributes such as toughness or friendliness have enjoyed
mixed success \cite{labov2001}. Our model can explain why high degree or high status people
eventually adopt grassroots linguistic innovations without recourse to a covert value system.
Overall, our results support claims that social solidarity, and not differences in status, is the
dominant factor in the adoption of linguistic innovations \cite{milroy, eckert}.

We saw that categorical decisions in the hubs scenario permit a well-connected group of people to
effect a rapid and complete change that it favors. Less well-connected people may be initially
biased against the change, but are swept along. This scenario does not appear to be relevant
for linguistic changes. Is the  model appropriate for some other type of
change in social norms or collective opinions? A categorical decision function maximizes an
individual's statistical expectations \cite{friedman} in the case of perfect knowledge. This low-temperature decision rule is
normally used in economic models \cite{friedman}, a reflex of economists' assumption that
individuals have complete information and  make independent and rational decisions to maximize
their gains. A recent example of a shift in public opinion in the economic domain was the 1999
repeal of Glass-Steagall Act provisions that separated investment banking from commercial banking.
Our model shows how a well-connected network of investment bankers who were biased in favor of
opportunities for investment banking could cause this change to be widely accepted (even if other
people held the opposite bias). More generally, a categorical decision rule is adaptive if the
utility of the competing options is well-defined and generally known, or if the cost of waiting for more
information is great. In times of revolution or
war, it may be clear that people who delay or vacillate in their allegiances are less safe than
people who take a side. Insofar as political behavior reflects this perception of the  risks, our model predicts 
people would apply a highly categorical decision process, be strongly influenced by a slight advantage of one option
over the other, and that as a result public
opinion in a time of crisis could be effectively controlled by a well-connected minority with shared
biases. Insofar as political behavior strikes a balance between taking a stand and waiting for more information,
our model suggests that grassroots political changes could take place without any definitive evidence about 
their ultimate utility.


\section*{Acknowledgments}

We thank L.A.N Amaral for time on his computing cluster. We are also grateful to P.J. Lamberson and Daniel Lassiter
for useful discussions about decision functions, and to Simon Todd for  incisive feedback on the mathematical exposition.

We are also grateful for financial support from JSMF Grant No. 21002061  and John Templeton Foundation Award No. 36617 . The opinions expressed in this publication 
are those of the authors and do not necessarily reflect the views of the John Templeton Foundation.

\bibliography{grassroots}

\newpage

\end{document}